\documentclass[pmlr]{jmlr}

\usepackage{longtable}

\usepackage{booktabs}
\usepackage[load-configurations=version-1]{siunitx} 

 \usepackage{tikz}

\theorembodyfont{\upshape}
\theoremheaderfont{\scshape}
\theorempostheader{:}
\theoremsep{\newline}

\jmlrvolume{204}
\jmlryear{2023}
\jmlrworkshop{Conformal and Probabilistic Prediction with Applications}
\jmlrproceedings{PMLR}{Proceedings of Machine Learning Research}

\title[Conformal Regression in Calorie Prediction for Team Jumbo-Visma]{Conformal Regression in Calorie Prediction for Team Jumbo-Visma}
\editor{Harris Papadopoulos, Khuong An Nguyen, Henrik Boström and Lars Carlsson}

  \author{\Name{Kristian {van Kuijk}} \Email{kristian.vankuijk@student.maastrichtuniversity.nl} \\ 
 \addr{Department of Advanced Computing Sciences, Maastricht University, The Netherlands} \\
 \addr{Visma Connect, The Hague, The Netherlands} \\ 
   \Name{Mark Dirksen} \Email{mark.dirksen@visma.com}\\
   \addr{Visma Connect, The Hague, The Netherlands}\\
   \Name{Christof Seiler} \Email{christof.seiler@maastrichtuniversity.nl}\\
    \addr{Department of Advanced Computing Sciences, Maastricht University, The Netherlands}\\
     \addr{Mathematics Centre Maastricht, Maastricht University, The Netherlands}
   }

\begin{document}

\maketitle

\begin{abstract}
UCI WorldTour races, the premier men's elite road cycling tour, are grueling events that put physical fitness and endurance of riders to the test. The coaches of Team Jumbo-Visma have long been responsible for predicting the energy needs of each rider of the Dutch team for every race on the calendar. Those must be estimated to ensure riders have the energy and resources necessary to maintain a high level of performance throughout a race. This task, however, is both time-consuming and challenging, as it requires precise estimates of race speed and power output. Traditionally, the approach to predicting energy needs has relied on judgement and experience of coaches, but this method has its limitations and often leads to inaccurate predictions. In this paper, we propose a new, more effective approach to predicting energy needs for cycling races. By predicting the speed and power with regression models, we provide the coaches with calorie needs estimates for each individual rider per stage instantly. In addition, we compare methods to quantify uncertainty using conformal prediction. The empirical analysis of the jackknife+, jackknife-minmax, jackknife-minmax-after-bootstrap, CV+, CV-minmax, conformalized quantile regression, and inductive conformal prediction methods in conformal prediction reveals that all methods achieve valid prediction intervals. All but minmax-based methods also produce sufficiently narrow prediction intervals for decision-making. Furthermore, methods computing prediction intervals of fixed size produce tighter intervals for low significance values. Among the methods computing intervals of varying length across the input space, inductive conformal prediction computes narrower prediction intervals at larger significance level.

\end{abstract}

\section{Introduction}
Nutrition is a key part of the performance of a rider. Until 2020, the Dutch cycling team Jumbo-Visma, winner of the \textit{Tour de France 2022}, would start preparing their calorie estimates up to three weeks in advance to ensure they had adequate estimates per cyclist and per stage. This is a time-consuming task. To improve team performance, we built regression models to predict the calories burned by a rider without needing any human computation. The models use information like the stage profile, the body mass index of a cyclist, or the race tactics, but also unforeseen factors such as the weather conditions. Following our forecasts, the nutritionists and cooks prepare meals for each rider per race day using the Jumbo Foodcoach app. This automated process ensures riders are provided with their exact nutrition needs, leading to a considerable advantage on race days.\\
\indent 
Despite a significant improvement in calorie prediction from the manual predictions of coaches ($R^2$ score of 0.55 for the prediction by coaches to 0.82 for the regression models), coaches still tune the output predictions. This means coaches tend to increase or decrease the models' outputs based on knowledge and previous experiences for specific races. Given this tendency for coaches to adjust the model predictions, instead of predicting a single outcome, it would be more beneficial to predict a range of possibilities. This can be achieved through prediction intervals. These intervals are calibrated based on the probability of encompassing the true output. By quantifying the reliability of the model predictions in estimating the speed and power of Team Jumbo-Visma riders, coaches can adapt predictions based on the uncertainty of the forecasts. To achieve this, we employ methods from the conformal prediction framework introduced by \cite{cp_creation}, providing valid and efficient prediction intervals. Each interval is computed given a significance value $\alpha$. This means if we take for instance $100$ Tour de France races and predict the calorie intake for a specific rider per race, in the long run, the true value will be outside the prediction bounds on average for only $\alpha$ races or less. Figure \ref{fig:illustration} illustrates our approach. After we compute prediction intervals for both the race speed and the rider's power output for a specific race, coaches combine both to obtain an energy forecast. As a concrete example, for one of the races of the 2022 season,  the long-term power forecast bounds were $[213, 265]$ for a specific rider, with a predicted power of $245.17$ (true value of $238.13$). Given the planned tactic and the previous experience of coaches with this race, the coach decided to round the power to $250$ watts. Combined with the predicted race time of $384$ minutes, computed from the speed forecast, this resulted in a calorie forecast of $5760$ kilocalories.\\
\tikzset{every picture/.style={line width=0.75pt}} 
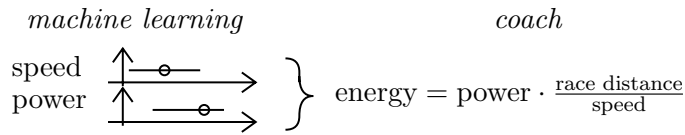
\begin{figure}
    \centering
\begin{tikzpicture}[x=0.75pt,y=0.75pt,yscale=-1,xscale=1]
\draw    (110.5,40) -- (146.5,40) ;
\draw    (122.5,60) -- (158.5,60) ;
\draw  (100,46.1) -- (174.5,46.1)(107.45,29) -- (107.45,48) (167.5,41.1) -- (174.5,46.1) -- (167.5,51.1) (102.45,36) -- (107.45,29) -- (112.45,36)  ;
\draw  (100,66.1) -- (174.5,66.1)(107.45,49) -- (107.45,68) (167.5,61.1) -- (174.5,66.1) -- (167.5,71.1) (102.45,56) -- (107.45,49) -- (112.45,56)  ;
\draw   (189.5,70) .. controls (194.17,70) and (196.5,67.67) .. (196.5,63) -- (196.5,63) .. controls (196.5,56.33) and (198.83,53) .. (203.5,53) .. controls (198.83,53) and (196.5,49.67) .. (196.5,43)(196.5,46) -- (196.5,41) .. controls (196.5,36.33) and (194.17,34) .. (189.5,34) ;
\draw   (126,39.94) .. controls (126.03,38.56) and (127.18,37.47) .. (128.56,37.5) .. controls (129.94,37.53) and (131.03,38.68) .. (131,40.06) .. controls (130.97,41.44) and (129.82,42.53) .. (128.44,42.5) .. controls (127.06,42.47) and (125.97,41.32) .. (126,39.94) -- cycle ;
\draw   (146,59.94) .. controls (146.03,58.56) and (147.18,57.47) .. (148.56,57.5) .. controls (149.94,57.53) and (151.03,58.68) .. (151,60.06) .. controls (150.97,61.44) and (149.82,62.53) .. (148.44,62.5) .. controls (147.06,62.47) and (145.97,61.32) .. (146,59.94) -- cycle ;

\draw (50,31) node [anchor=north west][inner sep=0.75pt]   [align=left] {speed};
\draw (50,51) node [anchor=north west][inner sep=0.75pt]   [align=left] {power};
\draw (213,41) node [anchor=north west][inner sep=0.75pt]   [align=left] {$\text{energy} = \text{power} \cdot \frac{\text{race distance}}{\text{speed}}$};
\draw (45,7) node [anchor=north west][inner sep=0.75pt]   [align=left] {\begin{minipage}[lt]{97.4pt}\setlength\topsep{0pt}
\begin{center}
 \ \ \textit{machine learning} \ \ 
\end{center}

\end{minipage}};
\draw (200,7) node [anchor=north west][inner sep=0.75pt]   [align=left] {\begin{minipage}[lt]{165.97pt}\setlength\topsep{0pt}
\begin{center}
\textit{coach}
\end{center}

\end{minipage}};

\end{tikzpicture}

\caption{Energy forecast procedure for Team Jumbo-Visma coaches. Machine learning provides prediction intervals. Coaches pick a value from the speed and power intervals and forecast energy consumption.}
\label{fig:illustration}
\end{figure}
\indent
In 2018, \cite{MLforcyclists} predicted the heart-rate response of a cyclist during a training session. The promising results led to a number of papers to predict the power performance of professional riders at the Tour de France (\cite{power_pred}), to predict the winner of the Tour de France (\cite{TDFwin}), to identify the next top cyclist (\cite{talent_identification}), and to athlete monitoring (\cite{tjv_athlete_monitoring}). Nevertheless, none of those methods quantify uncertainty in their predictions. The data science team of Visma Connect started working with Team Jumbo-Visma coaches and nutritionists in 2020 to improve the performance of the team using machine learning and mathematical methods. Previously, the calorie intakes were computed manually by the coaches using only domain knowledge from previous similar races and experience. But out on the track, unforeseen factors impact how much energy the cyclists burn. The weather, for instance, can cause cyclists to exert themselves more, or perhaps the tactics of the team needs to change due to other circumstances. This means coaches would often have to review their estimates several times before each stage of the race, a time-consuming exercise that had to be done for each rider for all races of the season.\\
\indent
We present the data that we received from Team Jumba-Visma in Section~\ref{data}. We introduce our baseline prediction model and review current conformal prediction methods in Section~\ref{cr-method}. We benchmark conformal methods on data from the Giro d'Italia and the Tour de France in Section~\ref{exp-res}. Finally, we interpret our findings and give recommendations on how coaches can fine-tune conformal methods in practice in Section~\ref{discussion}. 

\section{Data}\label{data}
The dataset for this paper consists of 1446 instances, all from the Team Jumbo-Visma men's team. The data is provided by Team Jumbo-Visma through \textit{Smartabase Human Performance Platform and Athlete Management System} (Smartabase), a data management platform for professional sport organizations and collected through a Garmin device and a crank based power meter. This allows to record duration, heart rate, speed, distance, elevation gains, calories, power, and other variables, of all race and training sessions. The \href{https://www.procyclingstats.com/index.php}{ProCyclingStats API}, a database that references all professional cycling races, allows to filter the training instances from the race data and provides more accurate race information such as the actual distance of the race, since it can happen that a rider forgets to turn off his Garmin devices at the end of a stage. For this project, we only retrieve the race name, race date, distance, race type (whether it is a one day race, stage race or Grand Tour) and name of the rider. Lastly, the \href{https://darksky.net/}{DarkSky API} provides all the weather information for each race, from the temperature to the wind effect at every race kilometer. \\
\indent 
The speed and power datasets use $8$ and $10$ features, respectively, ranging from the race type (one day race, stage race, or Grand Tour), the stage profile with the ascent/descent and distance, the weather conditions with the temperature, humidity, negative wind-effect and rainfall, attributes of the riders (body mass index (BMI)), and the tactics with the role of each rider (helper, climber or leader). The BMI of the rider and race strategy are only taken into account for the power dataset. 
\\
\indent
To take into account the steepness of a stage, we compute an ascent relation variable by dividing the ascent with the descent coefficient. Concerning the weather data, we obtain the negative wind effect by computing the mean of the wind speed when the degree of the cyclist is against the wind. This is done with the GPS files provided by Team Jumbo-Visma and by inspecting the direction the rider faces every three seconds. Lastly, we compute the rainfall by the product of the precipitation intensity times the precipitation probability. \\

\section{Methods}\label{cr-method}
\subsection{Prediction Model}\label{cr-method-1}
The time and the power are the two predicted response variables we provide to Team Jumbo-Visma. We train a random forest model as the underlying regressor as it performed best for our dataset in terms of $R^2$ and root-mean-square error. The energy $E$ is then computed mathematically from the time $t$ and the power $p$: $E = t \cdot p = \frac{d}{s}\cdot p$, with $d$ the race distance and $s$ the speed. This allows the Team Jumbo-Visma coaches to better understand our predictions and tweak them according to what they believe could be the actual power of a specific rider and stage. Predicting the energy directly makes it harder for the coaches to understand the logic underneath a certain prediction. In fact, even before we started working with the Dutch cycling team, the coaches first predicted the speed and power to finally obtain the needed calories intake of a specific rider. The predicted speed is the same for all the riders for each stage, while the power differs per rider. Hence, we use no information concerning individual riders for the speed estimator. Considering the very small finish time difference among riders compared to the overall length of a stage, we focus on predicting the average race speed. \\
\indent
In practice, we perform forecasts daily for Team Jumbo-Visma. Naturally, considering only one weather forecast for a race that takes place in more than 10 days is suboptimal. To solve this uncertainty, we assign weights for short-term forecasting based on how many days in advance the forecast is produced. Thus, the weather has a more important influence in the predictions closer to the race day. For example, five days prior to a race, we assign a weight of 0.9 to the model with weather features, and 0.1 to the one without weather features. For a forecast 10 days preceding the start of the race, the weights are of (0.5, 0.5), respectively. The pair of weights always sum to $1$ and are based on research from the NASA Space Place team at NASA's Jet Propulsion Laboratory (\cite{weather}).\\

\subsection{Conformal Prediction Methods}\label{cr-method-2}
In this section, we introduce the conformal prediction methods that we benchmark for Team Jumbo-Visma. A theoretical description of the methods mentioned can be found in \cite{jack+,phd_nonconf,CQR,cp_creation}. \\
\indent
Conformal prediction, introduced by Vovk, Gammerman and Shafer in 2005, has been proven to provide valid output (\cite{cp_creation}), i.e.\ predicted sets for any fixed confidence level $1-\alpha$ will not cover the true response with frequency at most $\alpha$. Since 2005, the conformal prediction framework has been applied to many learning algorithms, from support vector machines (\cite{cp_svm}), $k$-nearest neighbors (\cite{cp_knn}), to ridge regression (\cite{cp_ridge}), among others. In addition to providing uncertainties quantification for prediction, one of the main research focuses recently has been on providing a coverage of $\geq1-\alpha$ while keeping a low computational training cost. For instance, the jackknife+ method has a training cost equal to the training set size $n$. They train $n$ leave-one-out models and provide rigorous coverage guarantees regardless of the distribution of the data entries for any algorithm that treats the training points symmetrically (\cite{jack+}).\\
\indent
We investigate the following conformal regression methods: jackknife and its variations (jackknife+, jackknife-minmax, jackknife+-after- bootstrap, and jackknife-afterbootstrap-minmax), cross-validation (CV) and its variations (CV+ and CV-minmax), conformalized quantile regression (CQR) and inductive conformal prediction (ICP). 
\\
\indent
The different jackknife methods are based on the creation of $n$ set of leave-one-out models, with $n$ the size of our training set. This means we train $n$ models omitting each time one entry on which we test the performance of the model. The prediction intervals are then constructed from the $(1-\alpha)n$th smallest value of the empirical distribution of the absolute residuals of these omitted data entries. For better coverage, the jackknife can be adapted to the jackknife-minmax using simply the minimal and maximal value of the leave-one-out predictions. While the different jackknife methods do not suffer from overfitting, all have a lage training time. This does not pose a problem in our case since the dataset used is relatively small. \\
\indent
Nevertheless, for larger datasets the same methods can be applied using cross-validation, the CV method, instead of leave-one-out. This means we calibrate the model on a proportion of data. The refined method of the jackknife can also be implemented to the base CV method. To reduce the training time, the jackknife+ method can also be adapted operating a bootstrap approach, using only the available bootstrapped samples. This results in reduced computational time and is usually as robust as the jackknife+ method. \\
\indent All prediction intervals outputted by the jackknife and CV methods (and their respected refined methods) have constant width for all features across the input space. This behaviour is suboptimal since we desire the conformal predictor to reflect the certainty at a given feature value (\cite{phd_nonconf}). This is the case of the CQR and ICP methods. \cite{CQR} introduced the CQR method combining conformal prediction and classical quantile regression, inheriting the advantages of both. CQR uses quantile regressors to estimate the prediction bounds. \\
\indent
Lastly, the ICP method computes a nonconformity score indicating how different the instance is compared to other instances of the training set, and produces a corresponding $p$-value (\cite{cp_creation}). Those $p$-values express the proportion of the same or less conforming instances than the considered example. The nonconformity also takes into account the difficulty of predicting the response based on how different the instance is from other instances in the training set. We compute it by training a $k$-nearest neighbours regressor. This regressor returns the difficulty of predicting the outcome by considering the error of the underlying model. In this paper, we use as nonconformity score the absolute difference between the expected value and the value observed from the underlying regression model.

\section{Experiments \& Results}\label{exp-res}
\begin{figure}[!ht]
    \centering
    \includegraphics[width=\textwidth]{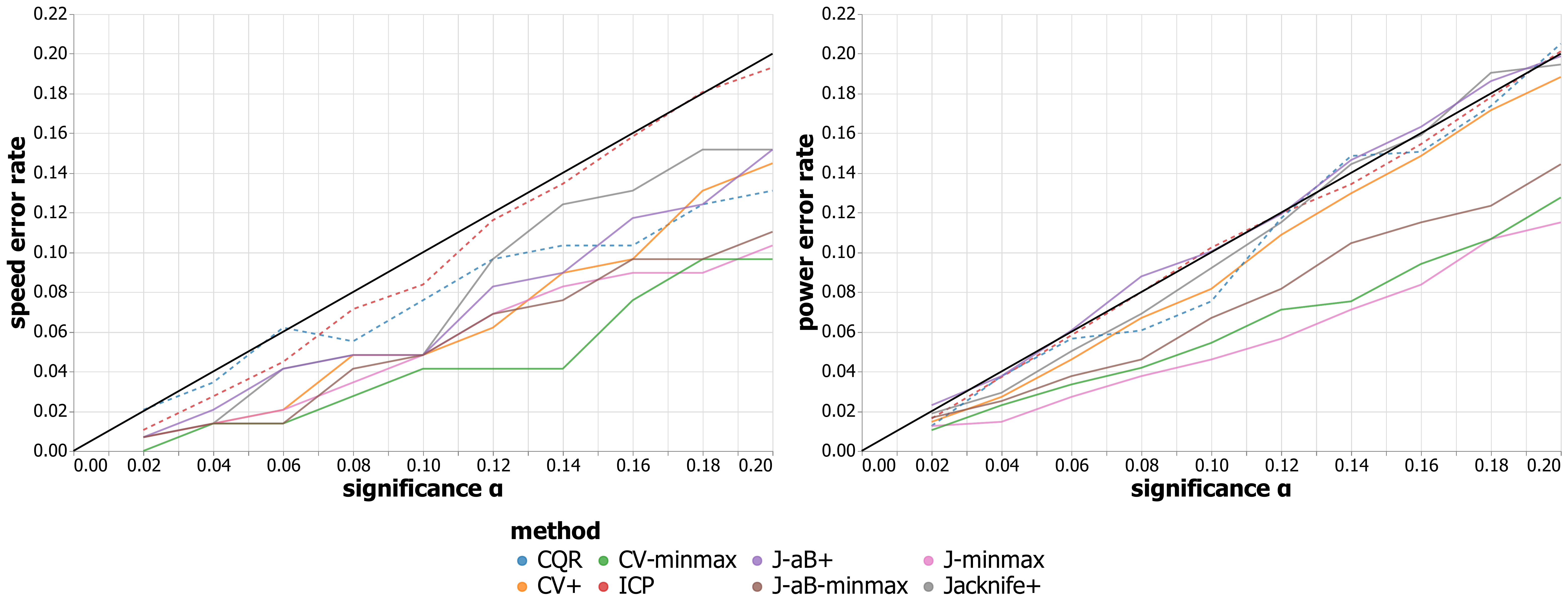}
    \caption{Error rate for the speed (left panel) and power (right panel) response variables (methods computing intervals of varying length across the input space are depicted by dashed lines).}
    \label{fig:error_rate}
\end{figure}

\begin{figure}[!ht]
    \centering
    \includegraphics[width=\textwidth]{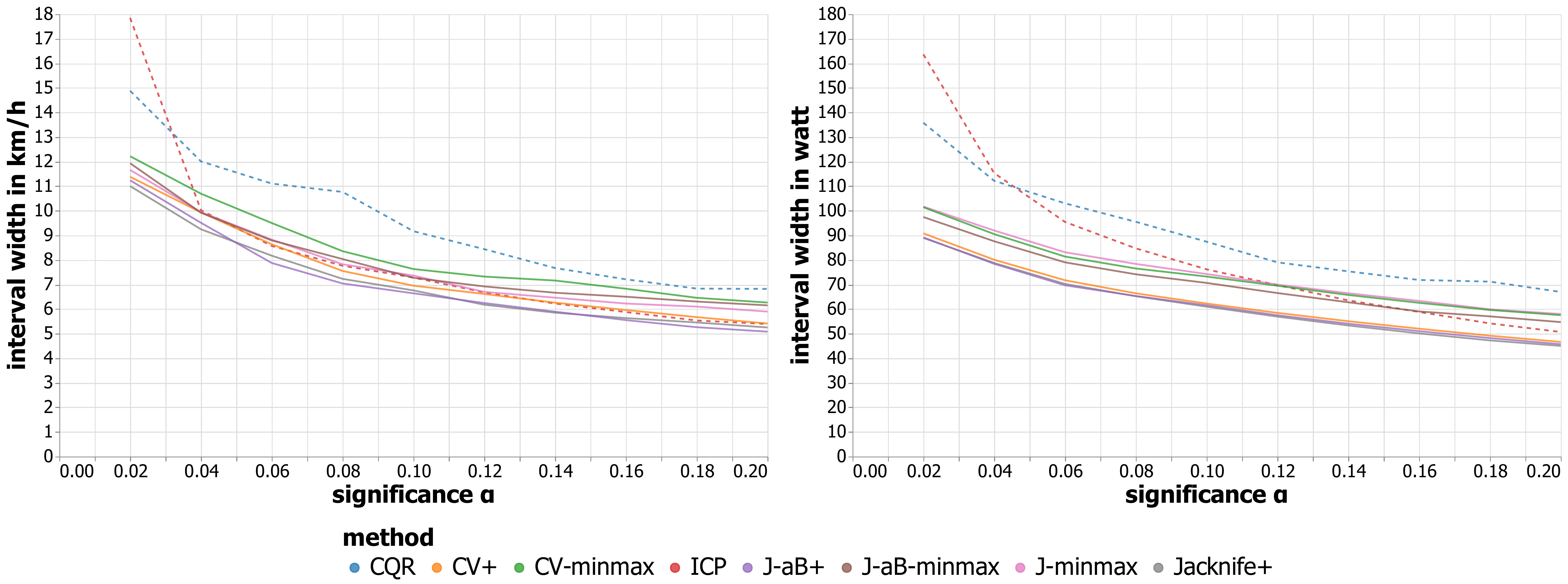}
    \caption{Interval width for the speed (left panel) and power (right panel) response variables (methods computing intervals of varying length across the input space are depicted by dashed lines).}
    \label{fig:interval_error}
\end{figure}

\begin{figure}[!ht]
    \centering
    \includegraphics[width=0.6\textwidth]{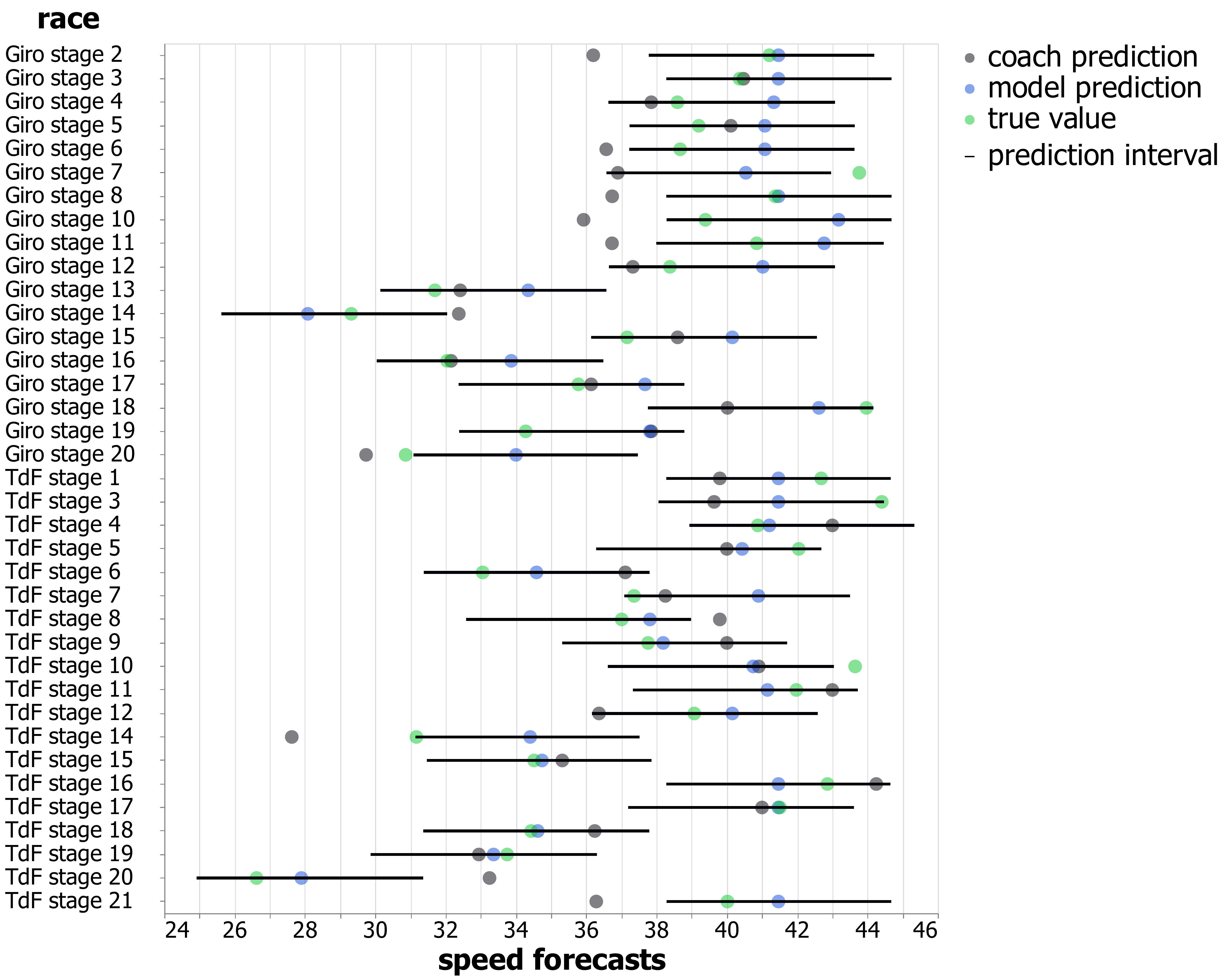}
    \caption{Speed forecasts comparison between coaches' manual predictions, regression models and prediction intervals (significance $\alpha=0.10$) to the true value for the Giro d'Italia 2019 and the Tour de France 2019 excluding time trials.}
    \label{fig:speed_comp}
\end{figure}

\begin{figure}[!ht]
    \centering
    \includegraphics[width=0.99\textwidth]{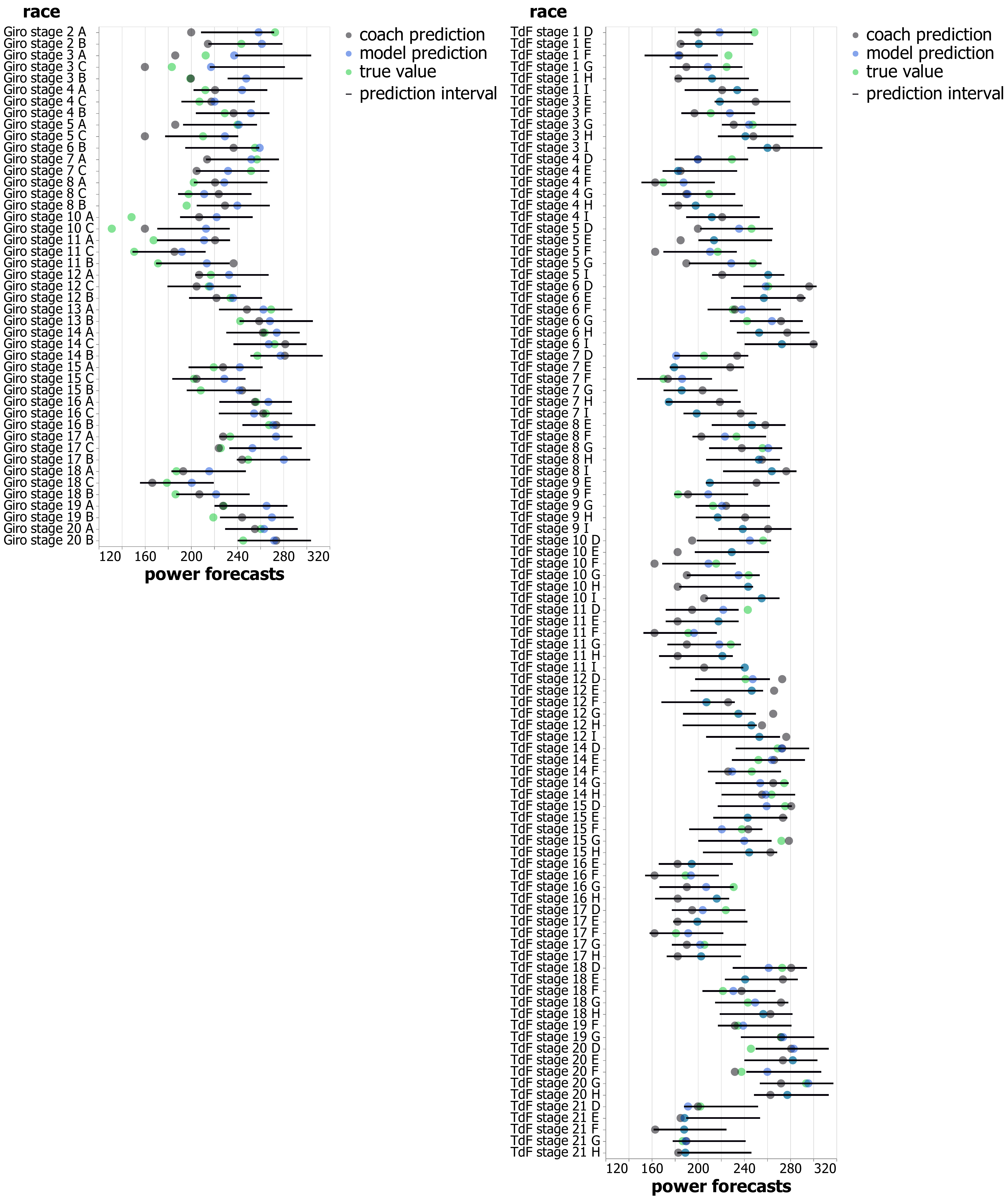}
    \caption{Power forecasts comparison between manual predictions of coaches, regression models and prediction intervals (significance $\alpha=0.10$) to the true value for the Giro d'Italia 2019 (for $3$ riders) and the Tour de France 2019 (for $6$ riders) excluding time trials (names of riders A-I are anonymized).}
    \label{fig:power_comp}
\end{figure}

For all the experiments, we preprocess the data and perform feature engineering (described in Section \ref{data}). We repeat five-fold cross-validation five times and report the average. We compare the error rate and interval width of all methods as a function of the significance $\alpha$ (Figures \ref{fig:error_rate} and \ref{fig:interval_error}) of the CV+, CV-minmax, jackknife+-after-bootstrap, jackknife-minmax-after-bootstrap, jackknife-minmax, jackknife+, CQR, and ICP methods for the speed and power response variables. To differentiate constant and non-constant interval size prediction intervals methods, the two methods computing non-constant interval size prediction intervals (CQR and ICP methods) are depicted by dashed lines. As significance levels larger than $0.20$ are very unusual, since the error rate becomes too large for the prediction intervals to be used in practice, all figures only include $\alpha \leq 0.20$. All experiments are performed on an Intel i7 with 8 CPU cores at 3GHz and 16GB of RAM. \\
\indent
The jackknife-minmax, jackknife-minmax-after-bootstrap and CV-minmax's error rates are considerably lower than the target value (Figure \ref{fig:error_rate}). The error rate at $\alpha=0.20$ is approximately $0.10$ for the three methods for the speed estimator. The jackknife+, jackknife+-after-bootstrap, CV+, ICP, and CQR methods compute prediction intervals close to the target value, particularly for the power response variable.\\
\indent
Concerning the interval widths (Figure \ref{fig:interval_error}), both response variables have a similar trend. The jackknife+, CV+ and jackknife+-after-bootstrap method produce the tightest intervals, particularly for the power response variable, followed by the minmax methods. The CQR and ICP methods produce considerably wider intervals for low $\alpha \leq 0.05$. Nevertheless, as $\alpha$ increases, the CQR and ICP methods intervals are comparable to the other methods.
\\
\indent
To illustrate the benefits of our approach, we compare in Figure \ref{fig:speed_comp} the manual predictions of coaches for two Grand Tour races of 2019 (Grand Tours are considered the most prestigious races of the season, typically spanning over three weeks) with our own forecasts, both single-point and prediction intervals, and the true response in predicting the race speed. An identical comparison for the power output can be found in Figure \ref{fig:power_comp}. We predict the race speed and the power output to obtain the energy needs per rider. The prediction of coaches tend to often fall outside the prediction intervals, while the error rate of our prediction interval is close to the target value. Furthermore, our model predictions demonstrate a greater degree of accuracy in predicting the true output compared to the predictions from coaches. The mean absolute error in Figure \ref{fig:speed_comp} is $2.46$ km/h for the coaches as opposed to $1.73$ km/h for the model ($\approx 30\%$ lower). For the power response, the mean absolute errors in Figure \ref{fig:power_comp} are $23.10$ watts and $12.68$ watts for the coaches and the model respectively ($\approx 45\%$ lower). We improve the $R^2$ for calorie prediction compared to manual predictions by $49\%$.


\section{Discussion}\label{discussion}
The jackknife+, jackknife+-after-bootstrap, and CV+ methods produce both valid and efficient prediction intervals for any $\alpha \leq 0.20$, with tight enough intervals to be useful in decision-making and the rate of instances outside of prediction intervals not exceeding $\alpha$.\\ \indent
The jackknife-minmax, jackknife-minmax-after-bootstrap and CV-minmax produce intervals with a lower error rate than the target value. This results from the commonality of those three methods using the minimal and maximal value of the leave-one-out or folds (\cite{jack+}). This behaviour is conservative considering we want the significance $\alpha$ to reflect the empirical error rate. The ICP and CQR methods are the only methods to reflect the certainty at a given feature value. They do not produce narrow enough intervals to be useful in decision-making for low $\alpha \leq 0.05$. Nevertheless, for larger $\alpha$, the produced intervals by the ICP method are comparable to the other methods.\\\indent
The CQR performs worst in terms of width of prediction intervals for the speed response variable, leading to wider intervals compared to other methods and an error rate under the target value $\alpha$. Our models are trained using a random forest model as underlying regressor. \cite{probCQR} showed that quantile regression forests are frequently excessively conservative, resulting in unnecessarily wide prediction intervals. ICP is computationally efficient as we only need to fit a single regression function. In contrast, we must run the regression function repeatedly when using the jackknife and CV approaches. These advantages come at a statistical price. If the training set size is much smaller than $n$, the size of the dataset, then the fitted model may be a poor fit, leading to wide prediction intervals. If instead, the training set size is close to $n$, then the calibration set is very small, leading to high variability (\cite{jack+}). ICP sacrifices different parts of the training set at different stages of prediction affecting its informational efficiency (\cite{cross-cp}). This may result in more conservative prediction intervals for our small dataset.\\
\indent
The minor differences in performance between the jackknife, CV (and their refined methods excluding minmax based methods), and ICP methods are not noticeable in decision-making according to Team Jumbo-Visma coaches. All methods compute valid prediction intervals that can be considered narrow enough by coaches to be useful. The ICP prediction intervals are comparable to the jackknife+ method for $\alpha \geq 0.05$ for the speed and $\alpha \geq 0.10$ for the power response variable. Most importantly, the latter produces prediction intervals of varying lengths across the input space. This means the ICP method reflects the certainty  of the model for a given feature value. Considering our small dataset, the heavy training cost of the different jackknife methods does not cause an issue ($1314$ seconds for the power response variable). \\
\indent
The power and speed data are different. We have power data for each rider, a total of $1446$ training instances. The power tends to differ greatly among riders in the same race. In contrast, we have speed aggregated data per race, a total of $436$ training instances. The smaller sample size for the speed response variable results in larger error rates. We believe that this affects the separation in error rate between minmax methods and other methods for the speed data. This effect is weaker for the power data due to the difference in the number of training instances.  \\\indent
Choosing a significance value $\alpha$ is an important part in the process of generating confidence intervals. The lower $\alpha$, the larger the prediction intervals. If the prediction sets are too wide, we risk that they are not useful anymore in decision-making. Low confidence results in high uncertainty about the true value. To pick the $\alpha$, we recommend a similar method to the elbow method applied to clustering (tracing back to \cite{elbow_clustering}). Figure \ref{fig:interval_error} suggests that both $\alpha = 0.04$ or $\alpha = 0.06$ are reasonable choices for the ICP method.\\

\section{Conclusion}
This paper introduces the calorie prediction project we started for Team Jumbo-Visma. Our energy forecasts are used daily by the coaches and nutritionists of the team. We provide Team Jumbo-Visma with prediction intervals that are narrow enough to be useful in practice. The ICP method performs best for our dataset. In future research, we plan to provide predictions for the women's team. We also plan to include domain knowledge through a Bayesian model, and to conformalize the posterior predictive intervals.\\

\bibliography{pmlr-sample}

\end{document}